\documentclass[10pt,twocolumn,letterpaper]{article}

\usepackage{iccv}
\usepackage{times}
\usepackage{epsfig}
\usepackage{graphicx}
\usepackage{amsmath}
\usepackage{amssymb}

\usepackage{enumitem}
\usepackage{makecell}
\usepackage{microtype}
\usepackage{subfigure}
\usepackage{booktabs} 
\usepackage{dblfloatfix}

\usepackage{mathtools}
\usepackage{amsthm}
\usepackage[symbol]{footmisc}

\usepackage[breaklinks=true,bookmarks=false]{hyperref}

\iccvfinalcopy 

\def\iccvPaperID{****} 

\ificcvfinal\fi

\begin{document}

\title{Painter: Teaching Auto-regressive Language Models to Draw Sketches}

\author{Reza Pourreza$^\dag$, Apratim Bhattacharyya$^\dag$, Sunny Panchal$^\ddag$, Mingu Lee$^\dag$, Pulkit Madan$^\ddag$, Roland Memisevic$^\ddag$ \\
Qualcomm AI Research\footnotemark[1]\\
$^\dag$San Diego CA, USA, $^\ddag$Markham ON, Canada\\
{\tt\small \{pourreza,aprabhat,sunnpanc,mingul,pmadan,rmemisev\}@qti.qualcomm.com}
}

\maketitle
\ificcvfinal\thispagestyle{empty}\fi

\begin{abstract}
Large language models (LLMs) have made tremendous progress in natural language understanding and they have also been successfully adopted in other domains such as computer vision, robotics, reinforcement learning, etc. In this work, we apply LLMs to image generation tasks by directly generating the virtual brush strokes to paint an image. We present \textit{Painter}, an LLM that can convert user prompts in text description format to sketches by generating the corresponding brush strokes in an auto-regressive way. We construct \textit{Painter} based on off-the-shelf LLM that is pre-trained on a large text corpus, by fine-tuning it on the new task while preserving language understanding capabilities. We create a dataset of diverse multi-object sketches paired with textual prompts that covers several object types and tasks. \textit{Painter} can generate sketches from text descriptions, remove objects from canvas, and detect and classify objects in sketches. Although this is an unprecedented pioneering work in using LLMs for auto-regressive image generation, the results are very encouraging.
\end{abstract}

\footnotetext[1]{ Qualcomm AI Research is an initiative of Qualcomm Technologies, Inc. }

\section{Introduction}
\label{intro}

Large language models (LLMs) are growing at an incredible pace and becoming ubiquitous solutions in every domain~\cite{gpt3, palm, flamingo, rt1, toolformer}. This tremendous success is partly owed to the auto-regressive nature of these models \emph{i.e.}, they look at the past and predict the future. 

Image generation and text-to-image conversion have seen a fast progress recently~\cite{stablediff, imagen, muse}. Current methods, despite very impressive results, are not explainable
. As such, it is hard to address the shortcomings of these methods.

Here, we present \textit{Painter}, an LLM that is employed for image generation. Unlike the existing image generation methods~\cite{stablediff, muse}, \textit{Painter} draws sketches the way humans do \emph{i.e.}, by generating a sequence of brush strokes in an auto-regressive way. Since this is an unprecedented work in this domain, we start with a relatively easier task which is sketch generation.

To train such a network, a dataset of text--image pairs is needed where the images should be expressed in the form of brush strokes. The only existing dataset in a similar format is Quick-Draw~\cite{quick_draw} which is a collection of 50 million class-label--drawing pairs across 345 categories. Since Quick-Draw always has a single object and there are no text descriptions, we create a new dataset by including single or multiple objects in a drawing, defining a composition or a relationship between them, and assigning a text prompt to each sample from a list of tasks. By training \textit{Painter} on the created dataset, it not only is able to draw sketches, it can also perform other tasks like completing incomplete sketches, wiping objects off a drawing, reproducing a sketch by generating the corresponding brush strokes and detecting and classifying the objects in a drawing.

Since the LLM used in \textit{Painter} needs to be multi-modal to consume interleaved text--image data, we make the necessary modifications to the vanilla LLM architecture to convert it to a language-vision model. This is done by adding residual cross-attention blocks that measure cross-attention between image features and hidden states of LLM. Furthermore, we equip the LLM with a visual feedback loop to monitor the state of the canvas as image generation progresses. This is similar to a robotics problem setting where the agent observes the state frequently.

Our contributions are as follows:
we introduce a model which to the best of our knowledge is the first to create images using auto-regressive language generation, we create a dataset of text-description--sketch pairs where the sketches are expressed in the format of strokes, and we enhance visual grounding in LLMs via feedback loop, cross-attention layers, and multitasking.

\section{\textit{Painter}}
\label{method}
In this section, we provide more technical details about the data generation process, model design, and the overall training method.

\subsection{Dataset}
To train \textit{Painter}, we need a dataset of text-description--sketch pairs where the sketches should be in strokes format \emph{i.e.}, all the brush movements should be recorded. To the best of our knowledge, Quick-Draw is the only large-scale dataset of this type. It consists of 50 million class-label--sketch pairs from 345 individual classes. There are two limitations in Quick-Draw for being directly used to train \textit{Painter}: 1) there is a single object in each sample. This could limit the capability of \textit{Painter} to learn the relationships between objects, object counts, and object compositions. 2) The text descriptions in Quick-Draw are limited to class labels. This provides a poor description of the objects in the sample and lacks concepts such as size and location. To alleviate these two limitations, we create the Multi-Object-Quick-Draw dataset. More details are provided in the next two subsections.

\subsubsection{Multi-object sketches}
Each sketch sample in Multi-Object-Quick-Draw contains one or more objects where the objects come from Quick-Draw and go through some processing. We establish an association between these objects by defining a relationship between them or adding relative location tags to them.

\textbf{Relationships}: to define a relationship between the objects, we follow Sketchforme's~\cite{sketchforme} approach with some modifications as follows. We initially parse the Visual Genome~\cite{visual_genome} relationships and select the relationships where both subject and object are present in Quick-Draw classes. Then we perform the followings processing per selected relationship, 1) normalize the bounding boxes of subject and object to our canvas size (256$\times$256) with a small random perturbation, 2) randomly select the subject and the object from Quick-Draw based on their associated classes, 3) scale their strokes to fit the normalized bounding boxes, and 4) put them on the canvas.
 
\textbf{Location tags}: the process above exhausts a small portion of Quick-Draw. For the remaining Quick-Draw samples, we divide them randomly into groups of 1, 2, 3, or 4 objects from the same or different classes and randomly place them across the canvas after the required size normalization.


\subsubsection{Text descriptions}
We define task-dependent text descriptions for the sketches in Multi-Object-Quick-Draw \emph{i.e.}, we associate a random task to the sketch from a list of predefined tasks and define a prompt based on the selected task.

\textbf{Tasks}: while the primary application of \textit{Painter} is text-to-sketch conversion, we train it on auxiliary tasks to improve the performance on the primary task via better object, location, and relationship grounding and adding complementary capabilities including wiping out object from a sketch and understanding the contents of a given sketch. Currently, the 6 following tasks are defined to create the dataset, however, this list can grow arbitrarily.
\begin{itemize}[noitemsep,topsep=0pt,leftmargin=0.3cm]
    \item \texttt{generate-all}: includes drawing a single or multiple objects on a blank canvas.
    \item \texttt{generate-partial}: includes completing a partial object or adding new objects to a sketch.
    \item \texttt{remove-all}: includes wiping off the object(s) of a sketch.
    \item \texttt{remove-partial}: includes removing an object from a multi-object sketch.
    \item \texttt{reproduce}: includes reproducing a given sketch by generating the strokes that form the sketch.
    \item \texttt{classification}: includes counting and classifying the objects of a sketch.
\end{itemize}

\begin{table*}[!t]
\caption{Default prompts per task for different scenarios.}
\label{tab:default_prompts}
\begin{center}
\begin{small}
\begin{tabular*}{\linewidth}{@{\extracolsep{\fill}} lll }
\hline
Task & Scenario & Default prompt \\
\hline
\sc{Generate-all}       & \sc{single w/o location}  & Draw a sketch of \texttt{$<$class-article$><$class-name$>$} \\
                        & \sc{single w/ location}   & Draw \texttt{$<$class-article$><$class-name$><$location-tag$>$} this sketch \\
                        & \sc{multi w/ rel}         & Draw a sketch of \texttt{$<$relationship-tag$>$} \\
                        & \sc{multi w/o rel}        & Draw a sketch of \texttt{$<$objects-list$>$} \\
\hline
\sc{Generate-partial}   & \sc{single}               & Complete this sketch as \texttt{$<$class-article$><$class-name$>$} \\
                        & \sc{multi w/ location}    & Add \texttt{$<$class-article$><$class-name$><$location-tag$>$} this sketch \\
                        & \sc{multi w/o location}   & Add \texttt{$<$class-article$><$class-name$>$} to this sketch \\
\hline
\sc{Remove-all}         & \sc{single}               & Remove \texttt{$<$class-article$><$class-name$>$} from this sketch \\
                        & \sc{multi}                & Remove all the objects from this sketch \\
\hline
\sc{Remove-partial}     & \sc{multi w/ location}    & Remove \texttt{$<$class-article$><$class-name$><$location-tag$>$} this sketch \\
                        & \sc{multi w/o location}   & Remove \texttt{$<$class-article$><$class-name$>$} from this sketch \\
\hline
\sc{Reproduce}          & \sc{all}                  & Reproduce this sketch \\
\hline
\sc{Classification}     & \sc{single}               & What is the class of this sketch \\
                        & \sc{multi w/ location}    & What is the object \texttt{$<$location-tag$>$} this sketch  \\
                        & \sc{multi w/o location}   & What are the objects in this sketch \\
\hline
\end{tabular*}
\end{small}
\begin{tiny}
\begin{itemize}[noitemsep]
\item[*] \texttt{$<$location-tag$>$} $\in$ \{\texttt{at the top of}, \texttt{at the bottom of}, \texttt{at the center of}, \texttt{at the right side of}, \texttt{at the left side of}, \texttt{at the top right corner of}, \texttt{at the top left corner of}, \texttt{at the bottom right corner of}, \texttt{at the bottom left corner of}\}. \\
\item[*] \texttt{$<$class-article$>$} $\in$ \{\texttt{a}, \texttt{an}, \texttt{the}\} depending on \texttt{$<$class-name$>$}. \\
\item[*] \texttt{$<$relationship-tag$>$} includes \texttt{subject}, \texttt{object}, and the \texttt{relationship} between them.\\ \item[*] \texttt{$<$objects-list$>$} refers to the objects in the sketch with their counts.\\
\item[*] \texttt{rel} refers to the cases where a relationship (from Visual Genome) exists between the objects.
\end{itemize}
\end{tiny}
\end{center}
\vskip -0.1in
\end{table*}

\begin{figure*}[!b]
\vskip 0.2in
\begin{center}
\centerline{\includegraphics[width=\textwidth]{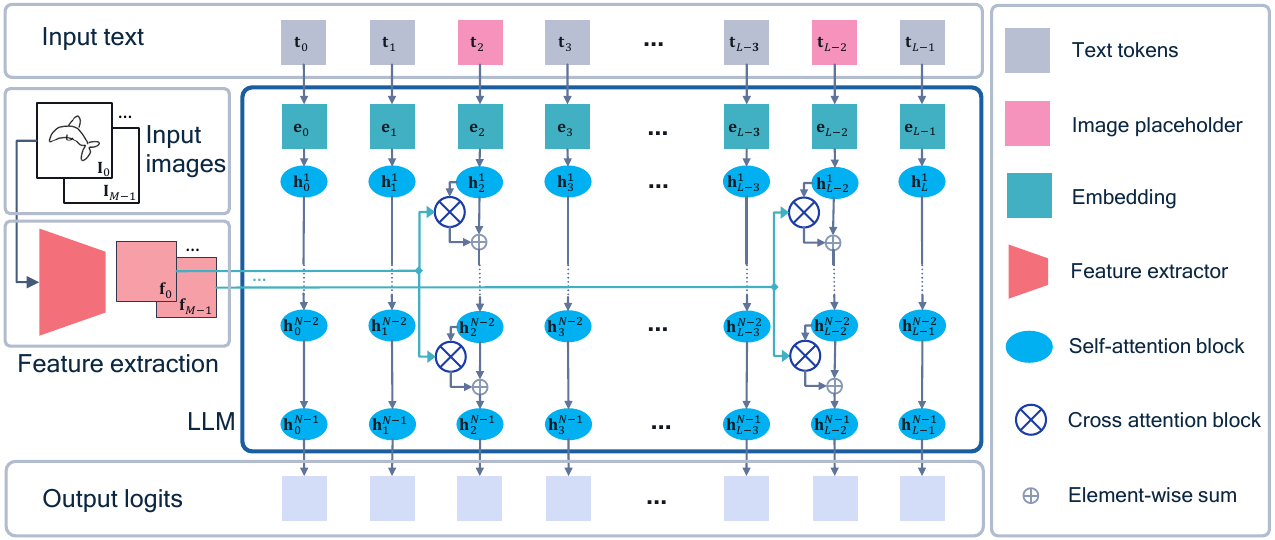}}
\caption{Block diagram of the multi-modal model.}
\label{fig:architecture}
\end{center}
\vskip -0.2in
\end{figure*}

\begin{table*}[!t]
\caption{Iterative inference steps.}
\label{tab:inference_steps}
\begin{center}
\begin{tiny}
\begin{tabular*}{\linewidth}{@{\extracolsep{\fill}} |l|l|l| }
\hline
Step & Prompt & Response \\
\hline
1   & \makecell[l] {\texttt{$<$command$>$ Draw an apple $<$image-placeholder$>$ $<$/command$>$} }
    & \makecell[l] {\texttt{$<$response$>$ $<$stroke$>$ color ... width ... points ... $<$/stroke$>$} }\\
\hline
2   & \makecell[l] {\texttt{$<$command$>$ Draw an apple $<$image-placeholder$>$ $<$/command$>$}  \\
                    \texttt{$<$response$>$ $<$stroke$>$ color ... width ... points ... $<$/stroke$>$} \\
                    \texttt{$<$image-placeholder$>$} }
    & \makecell[l] {\texttt{$<$stroke$>$ color ... width ... points ... $<$/stroke$>$} } \\
\hline
... & ... & ... \\
\hline
N-1 & \makecell[l] {\texttt{$<$command$>$ Draw an apple $<$image-placeholder$>$ $<$/command$>$}  \\
                    \texttt{$<$response$>$ $<$stroke$>$ color ... width ... points ... $<$/stroke$>$} \\
                    \texttt{$<$image-placeholder$>$} \\
                    ... \\
                    \texttt{$<$stroke$>$ color ... width ... points ... $<$/stroke$>$} \\
                    \texttt{$<$image-placeholder$>$} } 
    & \makecell[l] {\texttt{$<$stroke$>$ color ... width ... points ... $<$/stroke$>$} } \\
\hline
N   & \makecell[l] {\texttt{$<$command$>$ Draw an apple $<$image-placeholder$>$ $<$/command$>$}  \\
                    \texttt{$<$response$>$ $<$stroke$>$ color ... width ... points ... $<$/stroke$>$} \\
                    \texttt{$<$image-placeholder$>$} \\
                    ... \\
                    \texttt{$<$stroke$>$ color ... width ... points ... $<$/stroke$>$} \\
                    \texttt{$<$image-placeholder$>$} \\
                    \texttt{$<$stroke$>$ color ... width ... points ... $<$/stroke$>$} \\
                    \texttt{$<$image-placeholder$>$} } 
    & \makecell[l] {\texttt{$<$/response$>$} } \\
\hline
\end{tabular*}
\end{tiny}
\end{center}
\tiny{** The first \texttt{$<$image-placeholder$>$} in prompt corresponds to a blank canvas that is passed to the model to draw on and the subsequent ones correspond to canvas state feedbacks after each stroke.}
\vskip -0.1in
\end{table*}

\begin{table*}[!b]
\caption{Quantitative results.}
\label{tab:quantitative}
\begin{center}
\begin{small}
\begin{tabular*}{\linewidth}{@{\extracolsep{\fill}} llcccc }
\hline
Model & Train Dataset & \makecell[c] {\texttt{classification} \\ \texttt{accuracy (\%)} } 
                      & \makecell[c] {\texttt{remove-all}     \\ \texttt{PSNR (dB)} }
                      & \makecell[c] {\texttt{remove-partial} \\ \texttt{PSNR (dB)} }
                      & \makecell[c] {\texttt{reproduce}      \\ \texttt{PSNR (dB)} }\\
\hline
OPT-125M & \makecell[l]{ \sc{Multi-Object-Quick-Draw} }                    & 34.50 & 20.15 & 22.69 & 17.74 \\
\hline
OPT-125M & \makecell[l]{ \sc{Multi-Object-Quick-Draw} + \\ \sc{The Pile} } & 17.57 & 20.25 & 22.95 & 17.21 \\
\hline
OPT-1.3B & \makecell[l]{ \sc{Multi-Object-Quick-Draw} }                    &  3.20 & 19.56 & 22.27 & 17.20 \\
\hline
OPT-1.3B & \makecell[l]{ \sc{Multi-Object-Quick-Draw} + \\ \sc{The Pile} } & 40.26 & 20.64 & 23.90 & 16.97 \\
\hline
\end{tabular*}
\end{small}
\end{center}
\vskip -0.1in
\end{table*}

\textbf{Prompt text}:
For each sketch in Multi-Object-Quick-Draw, we randomly select a task. Then, depending on the task and the number of objects and the associations between the objects in the sketch, we define a prompt text. Table~\ref{tab:default_prompts} shows the default prompt texts for each task at different scenarios where the scenario can be the number of objects, whether a specific location is defined, whether a relationship exists between the objects, etc. In order to diversify prompt texts, we extend the default prompt texts by rephrasing them in several ways using a pre-trained BLOOM-176B model~\cite{bloom}. We randomly select one of the generated prompts to assign to a sketch. 

Once a task-dependent prompt is assigned to a sketch, a prompt sketch and a ground-truth sketch are generated where both could include modifications compared to the original sketch based on the selected task. The prompt sketch is given to the network as a part of the prompt and the ground truth sketch is used for supervised training. Task-dependent modifications can vary widely, for example in the \texttt{generate-all} task, all the objects are removed from the sketch to generate the prompt sketch and a blank canvas is given to the network to begin with, while the ground-truth sketch does not include any modifications.

\subsection{Model details}
This section provides more details about the \textit{Painter} model, including how prompts and responses are encoded, how \textit{Painter} digests interleaved texts and images, and how to equip \textit{Painter} with a visual feedback loop to monitor the state of the canvas.

\textbf{Prompt format}: 
We use the HTML format to encode the prompts and the responses \emph{i.e.}, commands and responses enclosed between \texttt{$<$command$>$} and \texttt{$<$/command$>$} tags and \texttt{$<$response$>$} and \texttt{$<$/response$>$} tags, respectively. 
We encode strokes in a similar way and include stroke color and thickness as follows: 

\texttt{$<$stroke$>$ color R G B width W points x\textsubscript{1} y\textsubscript{1}, x\textsubscript{2} y\textsubscript{2}, ..., x\textsubscript{L} y\textsubscript{L} $<$/stroke$>$} 

where \texttt{x\textsubscript{i}} and \texttt{y\textsubscript{i}} are the coordinates of the points in a stroke in string format and \texttt{L} is the length of the stroke. Draw actions use black ink and thickness of 1 pixel, while remove actions redraw objects in white ink and thickness of 2 pixels.

To process interleaved texts and images in the model, we insert an image-placeholder in the text wherever an image is needed. As such, the full prompt text for the \texttt{classification} task becomes: 

\texttt{$<$command$>$ What is the class of this sketch $<$image-placeholder$>$ $<$/command$>$ $<$response$>$ A tree $<$/response$>$}

where the response section is used in training only. The same applies to other tasks.

\textbf{Multi-modal LLM}: 
We use an off-the-shelf pre-trained LLM and modify it similar to~\cite{LRR} to receive and process interleaved texts and images. The overall block diagram is shown in Figure~\ref{fig:architecture}. As can be seen there, images are fed to a separate head and are consumed via residual single-head cross-attention components \cite{DouKGZWL00LPGW22,abs-2211-13319} wherever there is a corresponding image placeholder in the text. To formalize the cross-attention behavior, let us assume $\mathbb{T} = \{ \mathbf{t}_0, \mathbf{t}_1, ..., \mathbf{t}_{L-1} \}$ and $\mathbb{I} = \{ \mathbf{I}_0, ..., \mathbf{I}_{M-1} \}$ denote the sequences of text tokens (including the image placeholder token) and images that are fed to the model. Here, $\mathnormal{L}$ and $\mathnormal{M}$ represent the lengths of text tokens and image sequences, respectively.

In the text input head of the LLM, the discrete text tokens $\mathbf{t}_i$ are converted to continuous embeddings $\mathbf{e}_i$ and passed through the self-attention blocks of the LLM to generate the hidden states $\mathbf{h}_i^l$, where both $\mathbf{e}$ and $\mathbf{h}$ belong to $\mathbb{R}^{L \times H}$, $\mathnormal{l}$ and $\mathnormal{H}$ represent the LLM layer index and the embedding/hidden-state dimensionality. In the image input head of the LLM, the images $\mathbf{I}_i$ are converted to features $\mathbf{f}_i \in \mathbb{R}^{M \times F} $ where $\mathnormal{F}$ represents image features dimensionality after flattening.

Where there is an image-placeholder in the text, let us assume location $\mathnormal{j}$, cross-attention between the corresponding image features $\mathbf{f}_i$ and the hidden states $\mathbf{h}_j^l$ are measured and added to the hidden states. In the cross-attention component, keys and values are extracted from $\mathbf{f}_i$ and queries are extracted from $\mathbf{h}_j^l$. This is done for all the LLM layers except the last one as formulated below:
\begin{equation}
  \mathbf{h}_j^l \mathrel{+}= \operatorname{Cross-Attn}(\mathbf{f}_i, \mathbf{h}_j^l),\ \operatorname{for}\ 0 \le \mathnormal{l} < L-1\ 
\end{equation}

It is worth noting that we concatenate positional embeddings to image features $\mathbf{f}_i$ before passing them to cross-attention blocks, to preserve spatial information.

\begin{figure*}[!t]
\vskip 0.2in
\begin{center}
\centerline{\includegraphics[width=\textwidth]{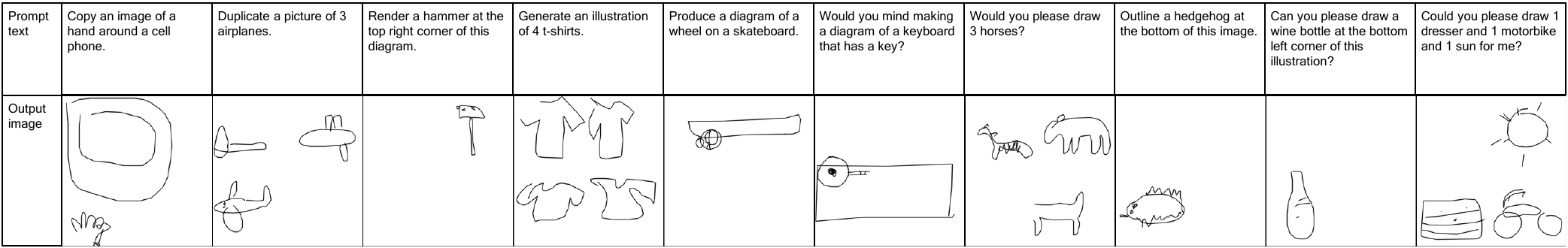}}
\tiny{(a)}
\centerline{\includegraphics[width=\textwidth]{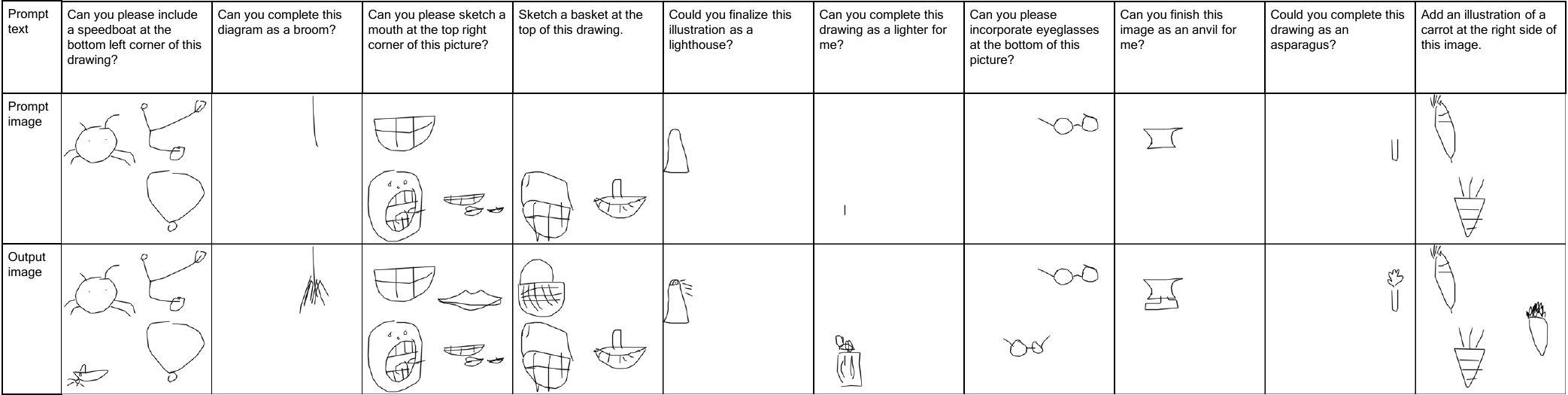}}
\tiny{(b)}
\caption{Selected (a) \texttt{generate-all} and (b) \texttt{generate-partial} results.}
\label{fig:generate}
\end{center}
\vskip -0.1in
\end{figure*}

\begin{figure*}[!b]
\vskip 0.3in
\begin{center}
\centerline{\includegraphics[width=\textwidth]{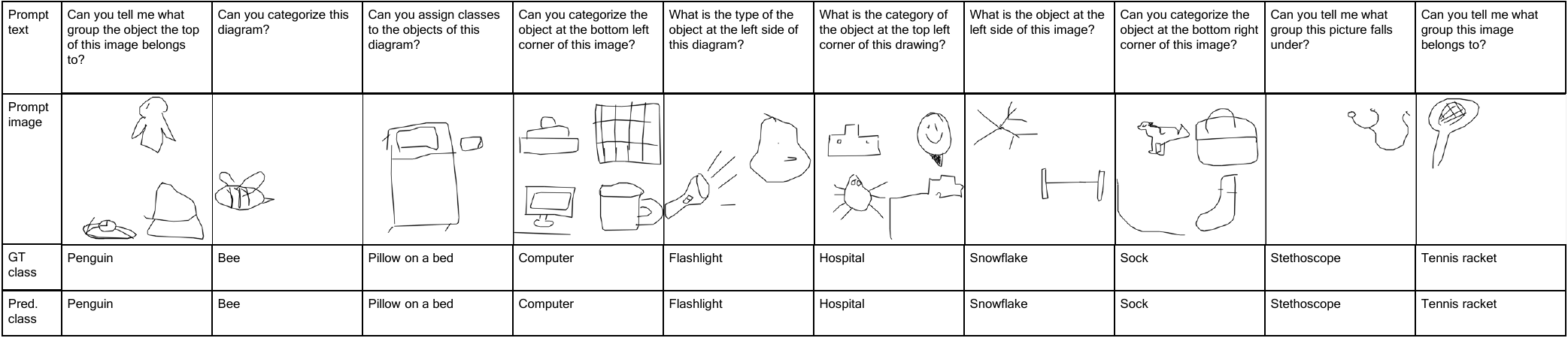}}
\caption{Selected \texttt{classification} results.}
\label{fig:classification_good}
\end{center}
\vskip -0.2in
\end{figure*}

\textbf{Visual feedback}: 
In order to mimic the way humans paint by looking at the canvas while drawing, we equip \textit{Painter} with a visual feedback loop to monitor the state of the canvas while generating strokes. As such, the generated strokes are applied on the canvas on-the-fly using an off-the-shelf line drawer. 
During training, prompts are augmented with the intermediate canvas states. During inference, recursive prompting is done \emph{i.e.}, once a stroke is generated by the LLM, is it drawn on the canvas, the feedback is applied, and a subsequent prompt with the updated prompt is executed. This is summarized in table~\ref{tab:inference_steps} for the \texttt{generate-all} task.

\subsection{Loss}
We finetune the LLM used in our model with a standard masked cross-entropy loss function via supervised training. We measure loss in the token domain on the response section only while the image placeholders are masked out. It is worth mentioning that we use the default LLM tokenizer's vocabulary as all the prompt contents are in string format and we do not introduce any special tokens.

\section{Experiments}
\label{experiments}

\begin{figure}[!t]
\vskip 0.2in
\begin{center}
\centerline{\includegraphics[width=\linewidth]{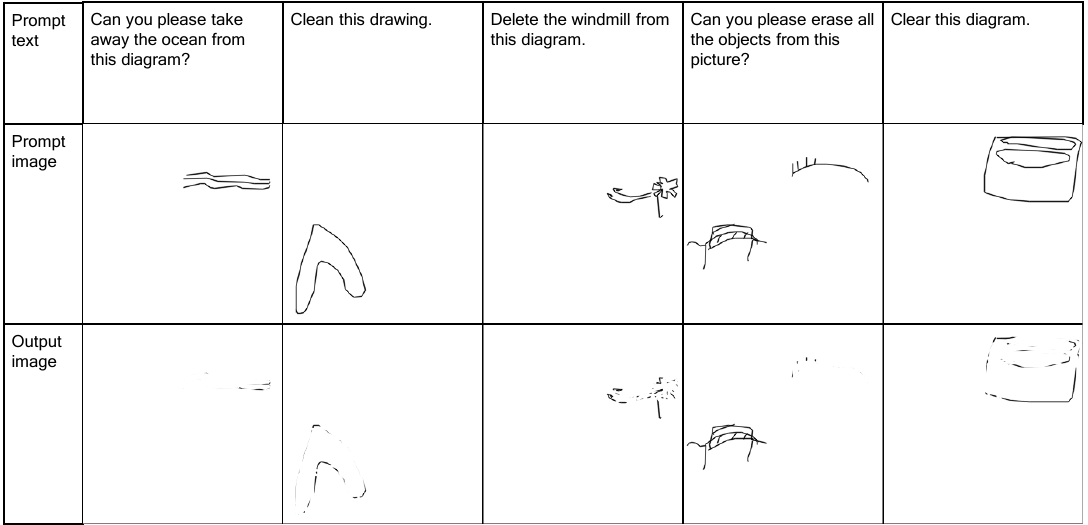}}
\tiny{(a)}
\centerline{\includegraphics[width=\linewidth]{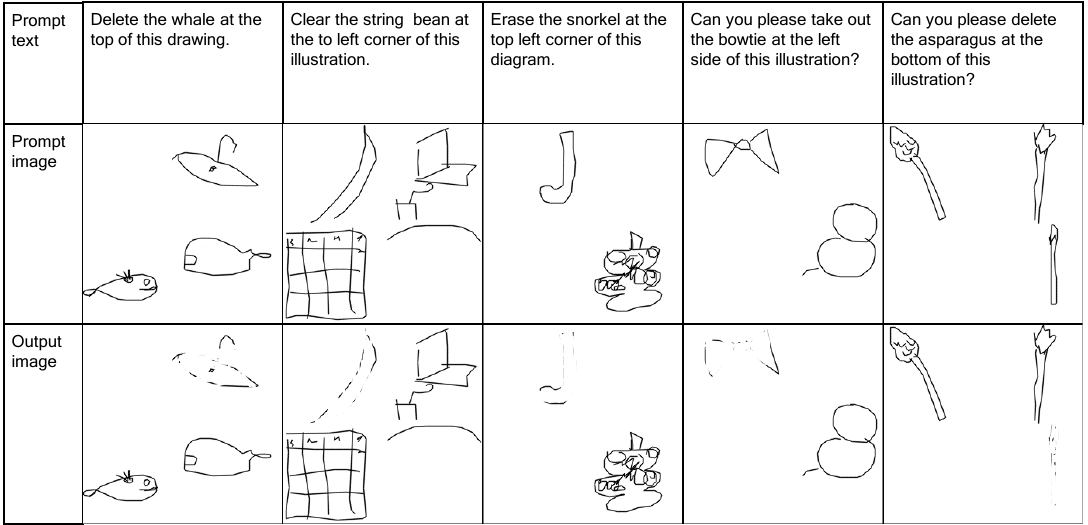}}
\tiny{(b)}
\centerline{\includegraphics[width=\linewidth]{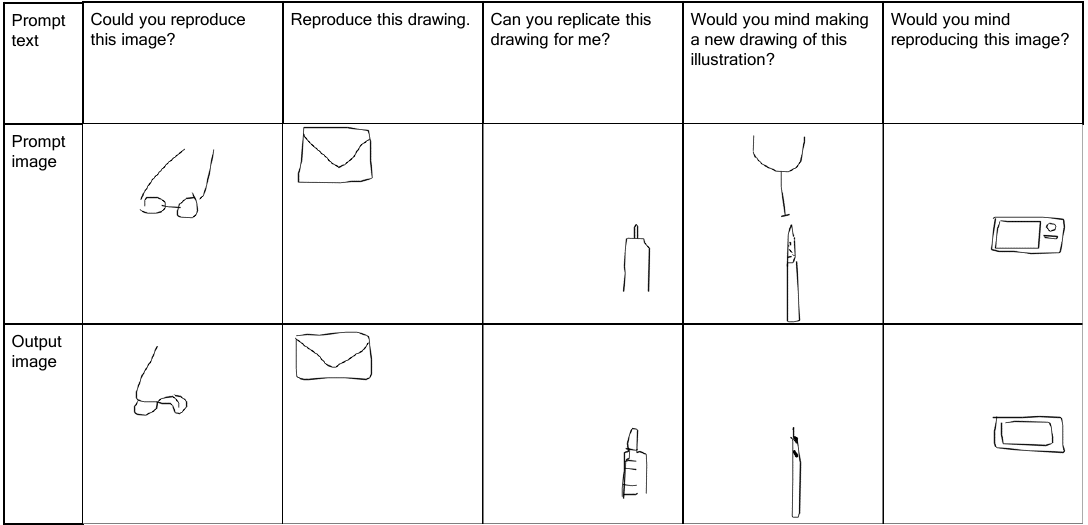}}
\tiny{(c)}
\caption{Selected (a) \texttt{remove-all}, (b) \texttt{remove-partial}, and (c) \texttt{reproduce} results.}
\label{fig:remove}
\end{center}
\vskip -0.2in
\end{figure}

\begin{figure*}[!t]
\vskip 0.2in
\begin{center}
\centerline{\includegraphics[width=\textwidth]{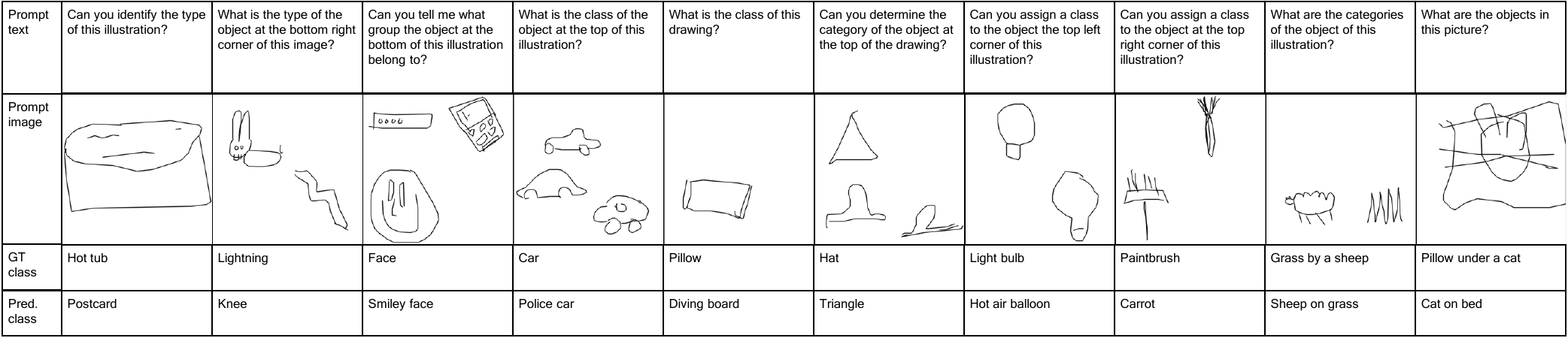}}
\caption{Selected incorrect \texttt{classification} results.}
\label{fig:classification_bad}
\end{center}
\vskip -0.1in
\end{figure*}

\subsection{Training setup}
The multi-modal architecture in \textit{Painter} is general enough to work with almost any off-the-shelf LLM. Here, we use two pre-trained LLMs from the OPT family~\cite{opt}, specifically OPT-125M and OPT-1.3B. Most vision-language models that are used in high-level vision tasks such as visual QA and image/video captioning~\cite{flamingo, palme}, use a pre-trained ViT~\cite{vit} or a CLIP~\cite{clip} image encoder to extract visual features. However, these image feature extractors are highly biased toward high-level visual features, while in \textit{Painter}, pixel-level visual information is required for better grounding and accurate guidance of the network. Hence, we use a pre-trained ResNet-50~\cite{resnet} for image feature extraction.

During training, the LLM is fine-tuned, the feature extractor remains frozen, and the randomly initialized cross-attention layers are trained. We train the models on Multi-Object-Quick-Draw. Multi-Object-Quick-Draw contains about 20 million samples, but here we use a subset of it with around 2M samples including all the samples with a relationship from Visual Genome (around 300K samples) and 1.7M samples with location tags. We use a small portion of this dataset containing around 1000 samples for evaluation and reserve the rest for training and validation. Since the LLM is already pre-trained on a large text corpus and is highly skilled in natural language understanding, we regularize the training on The Pile~\cite{pile} dataset to retain the natural language understanding capabilities of the LLM.

We train the models using Adam~\cite{adam} optimizer with a learning rate of $1e-5$ on two A100 Nvidia cards for one training epoch with a batch size of 4. 

During inference, we use greedy sampling in the \texttt{classification} task and use top-p sampling with $p=0.9$ for the other tasks.

\subsection{Results}
We can evaluate the tasks that \textit{Painter} is trained on as follows:
\begin{itemize}[noitemsep,topsep=0pt,leftmargin=0.3cm]
    \item \texttt{remove-all}, \texttt{remove-partial}, and \texttt{reproduce} can be evaluated using pixel-level metrics such as \texttt{MSE} and \texttt{PSNR}, since exact results are expected from them.
    \item \texttt{classification} accuracy can be measured by comparing the detected classes with the ground-truth classes.
    \item \texttt{generate-all} and \texttt{generate-partial} are not trivial for quantitative evaluation and need a user-study to rate the generated results. Since an extensive user-study is out of budget of this work, for these two tasks we show qualitative results only.
\end{itemize}

We report quantitative results for \texttt{classification} in terms of classification accuracy and for \texttt{remove-all}, \texttt{remove-partial}, and \texttt{reproduce} tasks in terms of \texttt{PSNR} in table~\ref{tab:quantitative}.

We show selected qualitative results for the \texttt{generate-all} and \texttt{generate-partial} tasks in figure~\ref{fig:generate}. The qualitative results that are shown in this paper are all generated using the \textit{Painter} model based on OPT-1.3B which is trained on Multi-Object-Quick-Draw and regularized on The Pile, unless otherwise noted. As can be seen from this figure, the model understands concepts such as shapes, relevant locations among objects, objects counts, and relationships.

Additionally, we show selected qualitative results for the \texttt{classification} task in figure~\ref{fig:classification_good} and for the \texttt{remove-all}, \texttt{remove-partial}, and \texttt{reproduce} tasks in figure~\ref{fig:remove}.

\subsection{Discussion}
As can be noted in table~\ref{tab:quantitative}, the classification accuracy is quite low. Our investigation leads to two main causes for this. The first reason is that we count a classification label as correct if there is an exact match between the ground truth label and the label that \textit{Painter} generates. There are examples such as ''pillow under a cat'' versus ''cat on a pillow'' that are conceptually the same, but are counted as wrong in our evaluation. The second reason is that there are objects classes that are very similar visually and are very hard to distinguish even for a human rater, such as ''birthday cake'' versus ''cake'' or ''pen'' versus ''marker''. Some examples from both categories of reasons are shown in figure~\ref{fig:classification_bad}.

\begin{figure*}[!b]
\vskip 0.1in
\begin{center}
\centerline{\includegraphics[width=\textwidth]{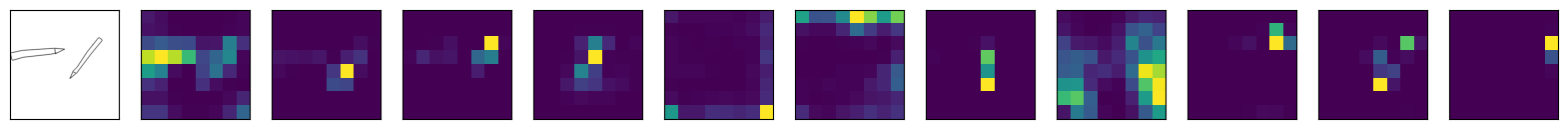}}
\tiny{(a) Prompt: What is the classification of the object at the right side of this image?}
\centerline{\includegraphics[width=\textwidth]{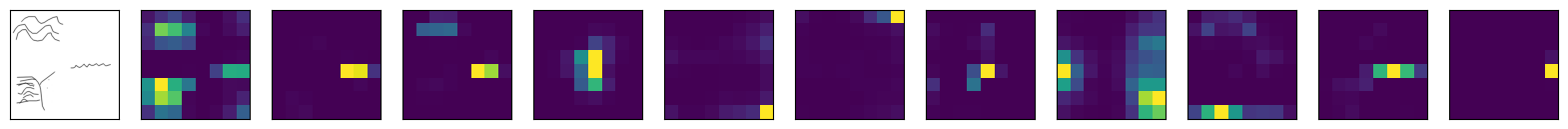}}
\tiny{(b) Prompt: What is the category of the object at the right side of this drawing?}
\centerline{\includegraphics[width=\textwidth]{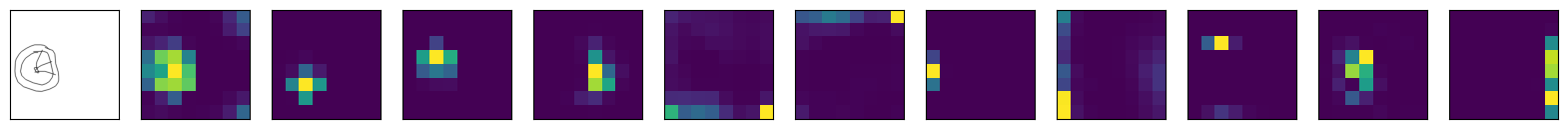}}
\tiny{(c) Prompt: What is the type of this picture?}
\centerline{\includegraphics[width=\textwidth]{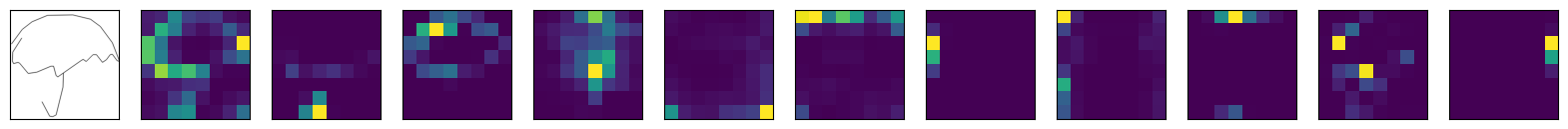}}
\tiny{(d) Prompt: Can you tell me what type of diagram this is?}
\centerline{\includegraphics[width=\textwidth]{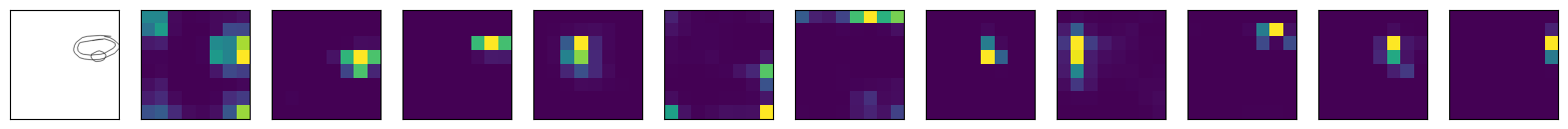}}
\tiny{(e) Prompt: Can you tell me the category of this drawing?}
\caption{Cross-attention maps of an OPT-125M-based \textit{Painter} model for several \texttt{classification} examples. There are 11 attention maps per example that from left to right correspond to the first layer to the eleventh layers in OPT-125M (there all cross-attention blocks in all 12 OPT-125M layers except the last one).}
\label{fig:attention}
\end{center}
\vskip -0.1in
\end{figure*}

To examine the attention maps that \textit{Painter} learns in the cross-attention blocks, we visualize them for a few \texttt{classification} examples in figure~\ref{fig:attention}. As can be seen here, \textit{Painter} can locate the objects in the input images and correctly attend to them, this is more pronounced in the first and third cross-attention blocks.

One of the shortcomings that \textit{Painter} has in its current form, is the limited number of object categories that it can identify which is limited to the 345 classes in Quick-Draw. We are working on expanding the object vocabulary in \textit{Painter} via techniques like reinforcement learning. 
\section{Related work}
\label{literature}
In spite of the great progress of auto-regressive LLMs and their extensive footprints in every domain, their full potential is not released in the image generation domain, as the latest methods utilize their representation learning aspect but do not benefit from their powerful auto-regressive nature. Therefore, we group existing image generation works into two broad categories: non-auto-regressive and auto-regressive methods.

\subsection{Non-auto-regressive methods}
Variational autoencoders (VAEs) \cite{vae, vae2} are one of the pioneering works for image generation. Later, Generative Adversarial Models (GANs) \cite{gan, stylegan, stylegan3} improved a lot upon VAEs on image generation and were considered the best-performing method, until recently. Diffusion models and their variants \cite{diffusion, stablediff, imagen, muse} are now state-of-the-art in this domain as they are very creative and generate mesmerizing results.

\subsection{Auto-regressive methods}
There are very a limited number of works in this category and to the best of our knowledge, none has a pre-trained auto-regressive LLM backbone for image generation. Sketch-RNN~\cite{sketchrnn} is an early work that generates sketches from a single class using a recurrent VAE. Sketchforme~\cite{sketchforme} extends to multi-object drawing via a combination of a transformer and a Sketch-RNN. Sketchformer~\cite{sketchformer} can classify, retrieve, and reconstruct sketches via an embedding that is learned in stroke space. Parti~\cite{parti} is a hybrid approach that uses an encoder-decoder LLM for language understanding and embedding and a pre-trained ViT-VQGAN~\cite{vitvqgan} for image generation.
\section*{Conclusions}
\label{conclusions}

We present \textit{Painter}, the first-ever LLM-based image generation solution that draws sketches by generating strokes in an auto-regressive way. We build the Multi-Object-Quick-Draw dataset consisting of diverse text-description--sketch pairs where the sketches contain single or multiple objects with relationships or relative location tags between the objects, and the text descriptions are devised from a list of pre-defined tasks with additional prompt diversification using a very large language model. We modify the LLM architecture by adding residual cross-attention layers to make it a vision-language model, and additionally add a visual feedback loop to actively observe the state of the canvas. Our results show the viability of \textit{Painter}'s approach.
There are shortcomings in \textit{Painter} including the limited number of object categories that will be addressed in future work.

{\small
\bibliographystyle{ieee_fullname}
\bibliography{bib}
}

\end{document}